\documentclass[acmsmall]{acmart}

\renewcommand\footnotetextcopyrightpermission[1]{} 
\AtBeginDocument{%
  }

\begin{document}

\title{Personalized Route Recommendation Based on User Habits for Vehicle Navigation}

\author{Yinuo Huang}
\authornotemark[1]
\email{huangyinuo@iem-toyota.cn}
\author{Xin Jin}
\affiliation{%
  \institution{Toyota Motor Engineering \& Manufacturing (China) Co., Ltd.}
  \city{Beijing}
  \country{China}
}

\author{Miao Fan}
\author{Xunwei Yang}
\author{Fangliang Jiang}
\affiliation{%
  \institution{NavInfo Co., Ltd.}
  \city{Beijing}
  \country{China}}

\renewcommand{\shortauthors}{Yinuo Huang et al.}

\begin{abstract}
  Navigation route recommendation is one of the important functions of intelligent transportation. However, users frequently deviate from recommended routes for various reasons, with personalization being a key problem in the field of research. This paper introduces a personalized route recommendation method based on user historical navigation data. First, we formulate route sorting as a pointwise problem based on a large set of pertinent features. Second, we construct route features and user profiles to establish a comprehensive feature dataset. Furthermore, we propose a Deep-Cross-Recurrent (DCR) learning model aimed at learning route sorting scores and offering customized route recommendations. This approach effectively captures recommended navigation routes and user preferences by integrating DCN-v2 and LSTM. In offline evaluations, our method compared with the minimum ETA (estimated time of arrival), LightGBM, and DCN-v2 indicated 8.72\%, 2.19\%, and 0.9\% reduction in the mean inconsistency rate respectively, demonstrating significant improvements in recommendation accuracy.
\end{abstract}




\maketitle

\section{Introduction}
\thispagestyle{empty}

Route recommendation plays a pivotal role in our daily lives as a crucial element of intelligent transportation systems. When a user requests route guidance during a trip, the navigation service generates multiple routes and selects the top three options best suited to the user's need. However, some users’ misalignment between the trajectory and the recommended route indicates deviations from the chosen route. This misalignment can stem from various factors, including the user's specific intentions, potential missing road data or traffic incidents along the recommended route, and the user's personalized route preferences. As shown in Fig. 1, the user deviates from the recommended navigation route and chooses a scenic route along the riverbank. Therefore, uncovering latent information within users' historical navigation data and tailoring personalized route recommendations can enhance user satisfaction with navigation services.
\begin{figure}[ht]
  \centering
  \includegraphics[width=0.5\linewidth]{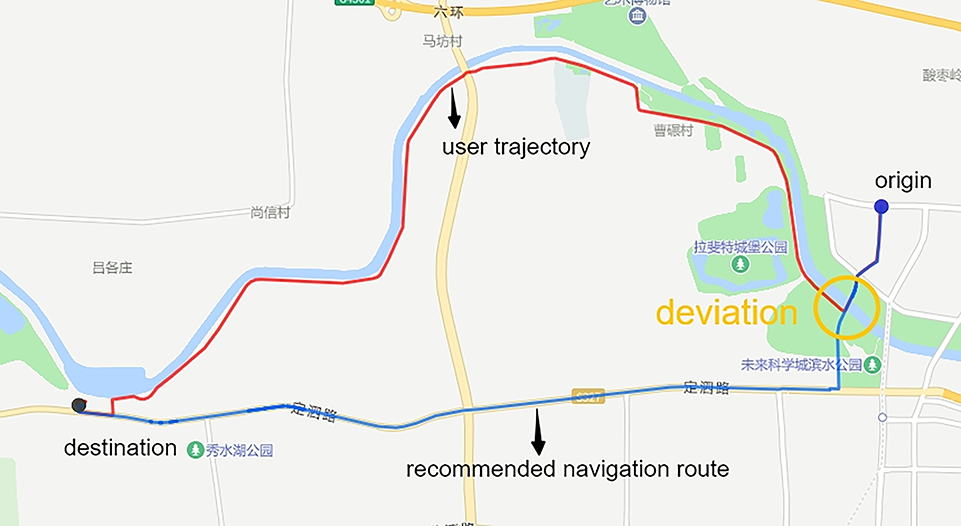}
  \caption{Misalignment between the user trajectory and navigation route. The user diverged from the navigation's recommended route, opting instead for a scenic route along the riverbank.}
  \Description{}
\end{figure}

With the rapid advancement of intelligent navigation services and the exponential growth of big data, navigation services preferred by most people often fall short of meeting users' diverse needs. Customized, personalized navigation has emerged as the prevailing trend. When selecting a recommended navigation route, users take into account various factors such as time, distance, traffic congestion, tolls, safety, and convenience. A route is composed of multiple links, and the attributes of each link and the road network encapsulate hidden information about the route. In addition to analyzing route information, learning user profiles is crucial for effective route recommendation.

To solve the above problems, a personalized route recommendation model based on user habits for vehicle navigation is presented in this research. In addition to the overall route information such as time, distance, toll, etc., link sequence features are introduced to learn the detailed information of the routes. Simultaneously, user profiles are extracted from historical trajectory data and navigation behaviors. Subsequently, a Deep-Cross-Recurrent (DCR) learning model is employed to facilitate personalized route recommendations. The principal contributions of this research are summarized as follows:
\begin{itemize}
\item Introduction of link sequence attribute features and landscape features to enhance the expression of route information.
\item Construct user profiles derived from historical trajectory data and navigation behavior to establish the groundwork for personalized route recommendation.
\item A DCR learning model is proposed for route recommendation. The model is tested on offline datasets and compared with other methods. The results validate the superior performance of the model.
\end{itemize}

\section{Related Work}
In the realm of navigation route research, Wuman Luo et al. leveraged trajectory data from a substantial user pool to construct mobile networks, identifying the most popular route between origins and destinations at different times \cite{Luo13}. This method effectively captures the preferences of the majority of users, it falls short of reflecting personalized preferences. Peisong Li et al introduce a road attribute description method by PS theory and a personalized route planning algorithm. After the user selects a travel preference plan, the priority of the route attribute is set, favoring routes that align closely with the target direction \cite{Li22}. However, it is difficult for users to accurately express their preferences for each criterion. 

In recent years, numerous scholars have employed deep learning methods to investigate personalized route recommendations. Jingyuan Wang et al. employed neural networks to learn the cost function within the A* algorithm for personalized route recommendations \cite{Wang19}. Ran Cheng et al proposed the R4 learning framework to predict the deviation rate of candidate routes, recommending a route with the lowest deviation rate to users \cite{Cheng21}. Shan Liu integrated the Dijkstra algorithm into deep inverse reinforcement learning to produce personalized routes when road network information is unknown \cite{Liu20}. Building upon this approach, Shan Liu et al introduced a graph attention network information and improved the IRL method to recommend personalized routes considering real-time traffic conditions \cite{Liu22}. However, when the state space is large, accurately computing the expected state visitation frequency becomes challenging. These studies inspire us to learn personalized route recommendations through deep learning techniques.

Route recommendation involves two key processes: recall and sorting. The recall phase aims to generate a wide array of relevant candidate routes between the origin and the destination. Subsequently, in the sorting phase, these recalled routes are ranked, with the top-ranked route being presented to the user. This paper primarily concentrates on route sorting, delving deeply into route data and user profiles derived from users' historical trajectories and navigation behaviors.

\section{Methodology}

In the context of navigation recommendation, the inconsistency rate indicates the degree of deviation from the route after the user selects navigation routes. It can be employed to evaluate the quality of route recommendations. A higher inconsistency rate is indicative of a user's dissatisfaction with the recommended route. In this paper, we define the personalized route recommendation problem as follows: Assuming there is a navigation database, given origin, destination, and request time, our objective is to predict the inconsistency rate of candidate routes and recommend the route calculation with the lowest inconsistency rate to the user. The inconsistency rate is defined as follows:
\begin{equation}
  IR=1-\frac{dis_{tc}}{dis_t}
\end{equation}
where \(IR\) indicates the inconsistency rate, \(dis_t\) denotes the overall distance of the user’s trajectory. By mapping trajectory data into the road network, we obtain the sequence of track links, the candidate route also be represented by the link sequence, \(dis_{tc}\) represents the sum length of same links between the user’s trajectory and the candidate route. If the user deviates from the route, the label IR is set to 0 and 1 otherwise.

\subsection{Feature Extraction}

With the widespread adoption of navigation services, a significant volume of user trajectory data is generated daily. To provide a more comprehensive representation of navigation recommendations, we first systematically build rich features derived from these datasets. We summarize the features into two aspects: route features and user profiles.

\subsubsection{Route features.}\ 

Spatial information: By mapping trajectory data into the road network, we can extract overall information about the route, including distance, number of traffic lights, tolls, number of turns, and so on. Additionally, we can gather data on the sequence of links, such as link length, number of lanes, and road types. Location-type information can be acquired through the POI grid vector.

Temporal information: The time of route recommendation requests is also a pivotal factor in navigation recommendations. This time can be refined into various features, including whether it falls within peak periods, weekends, specific days of the week, or hours of the day.

Traffic Information: The road conditions along the route significantly impact the user’s driving experience. Individuals in a hurry typically seek to avoid congested routes. Traffic information can be conveyed through various features, such as different levels of road conditions at the time of departure, et al. 

landscape Information: In current research, the incorporation of scenic elements along routes is often overlooked. Considering the significant cost involved in manual assessment, this paper suggests a methodological approach. As shown in Fig. 2, this paper initially processes the route data by gridding it after mapping it into the road network. Subsequently, it statistics the water system data and green spaces data of grid along the route. Such information can serve as a representation of the landscape attributes of the route.
\begin{figure}[ht]
  \centering
  \includegraphics[width=0.5\linewidth]{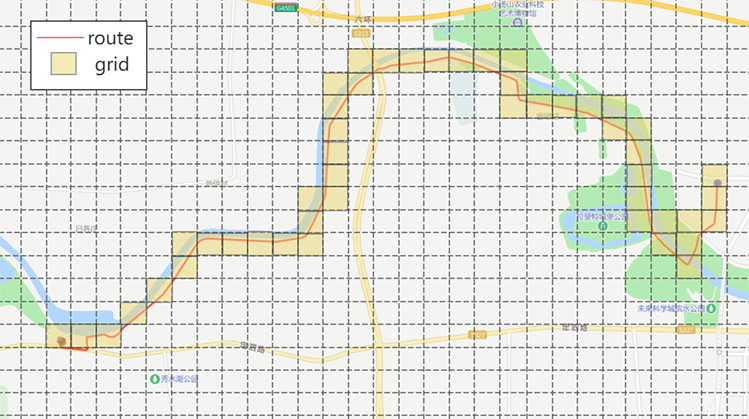}
  \caption{Landscape information extraction. Convert the route into a grid sequence and then statistic the number or proportion of grids that include water systems data and green spaces data.}
  \Description{}
\end{figure}

\subsubsection{User profiles.}\ 

The user’s personalized route preference refers to the relatively stable behavioral features of the user when navigating. Drawing from daily navigation experiences, along with the aforementioned route information and the user’s historical travel data, relevant behavioral characteristics are extracted. These include the user’s inconsistency ratio, the proportion of users opting for the fastest route, and so on. These characteristics help illuminate the user's personalized preferences, contributing to the establishment of a dataset for user profiles.

Cluster analysis is performed on the extracted user historical behavior features to distinguish the preference category of each user in the dataset. The K-Means algorithm is an unsupervised clustering technique. It partitions a given sample set into K clusters based on the distances between samples, aiming to tightly connect points within clusters while maximizing the separation between clusters. T-SNE is a nonlinear dimensionality reduction and data visualization technique that transforms high-dimensional data into two or three dimensions \cite{Maaten08}, preserving local relationships between data points to the greatest extent possible. As shown in Fig. 3, for a more intuitive analysis of user profiles based on historical behavioral data, this paper employs the K-Means algorithm to categorize users into 6 clusters and utilizes T-SNE for dimensionality reduction and visualization of the clustering outcomes.
\begin{figure}[ht]
  \centering
  \includegraphics[width=0.5\linewidth]{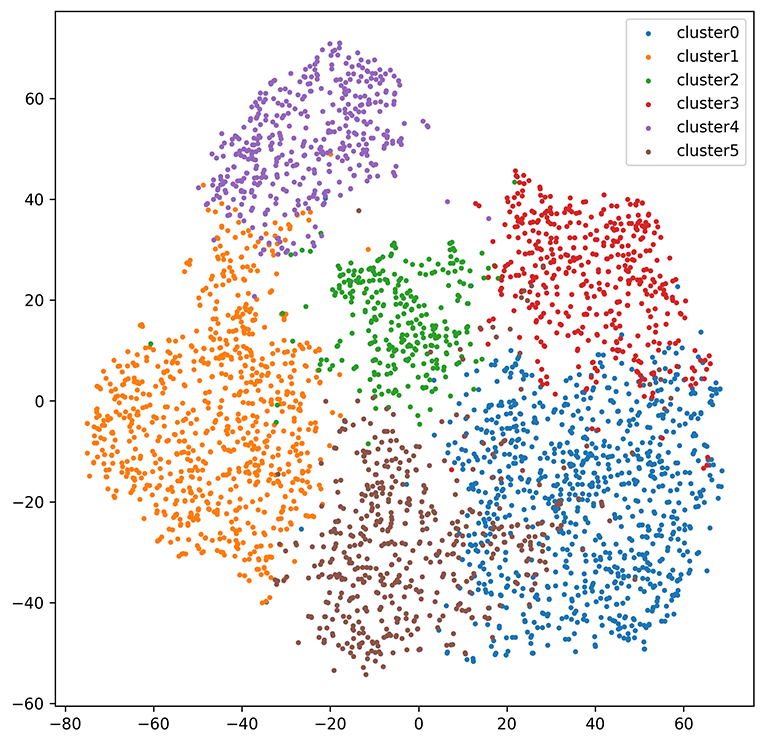}
  \caption{Visual cluster results of T-SNE. The K-Means method is used to cluster data and visualize them following dimensionality reduction by t-SNE.}
  \Description{}
\end{figure}

The cluster center of each category encapsulates comprehensive information about the cluster. A comparison of the features within each category reveals distinct differences in the feature variables, leading to the definition of corresponding labels for each category. Cluster 0 users are more toll-sensitive and willing to accept some time loss to choose routes with lower tolls. The user preference of cluster 1 is characterized by frequently choosing the fastest or shortest route. Users in cluster 2 prefer higher-quality roads, reflected in their preference for wider and safer routes. Cluster 3 users prioritize highways, often opting for highways when travel times are small, and they are insensitive to tolls. Users in cluster 4 prefer scenic routes, characterized by their selection of paths often near rivers or green parks. Cluster 5 users are sensitive to congestion and often opt for routes with smoother traffic conditions.

\subsection{Route Rank Model}

In route sorting applications, traditional machine learning methods like LightGBM have limitations in the hidden patterns within sequences of link attributes along the road. This paper proposes an advanced approach by integrating the DCN-v2 and LSTM models to construct a more complex model to solve the problem.

DCN-v2 is proposed for recommender systems \cite{Wang21}, an improved version of DCN. The cross-network introduces a mixture of expert network structures to enhance the cross-ability of different subspace features. The deep network comprises multiple layers of MLP designed to uncover underlying patterns effectively.
\begin{figure}[ht]
  \centering
  \includegraphics[width=0.5\linewidth]{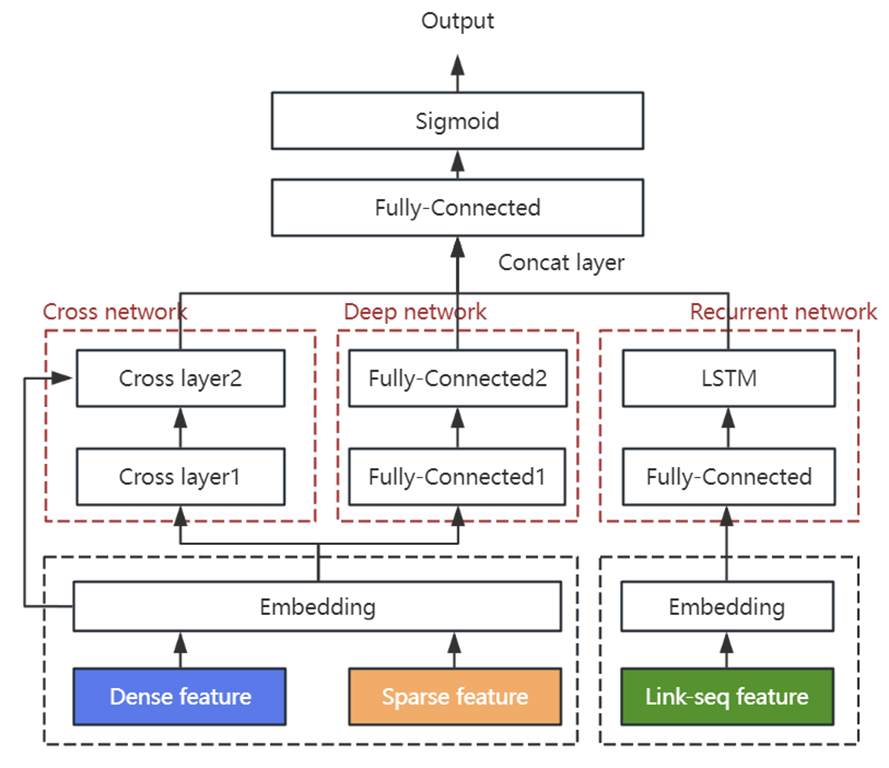}
  \caption{Structure of DCR. The DCN-v2 network is on the left and the recurrent network is on the right.}
  \Description{}
\end{figure}

LSTM is a special type of RNN \cite{Hochreiter1997}, that can learn long-term dependent information, and it has demonstrated excellent performance in natural language processing, speech recognition, and other domains. Similar to all recurrent networks, LSTM comprises a chain of repeating modules within a neural network. In each module of LSTM, the input gate, forget gate, output gate, modulated input, memory cell, and hidden state.

This paper combined DCN-v2 and LSTM models to build a Deep-Cross-Recurrent network for learning to estimate route sorting scores. The model structure is described in Fig. 4. The route features and user features mined above are taken as the input of the model. Initially, sparse features of the global route are converted into dense features via an embedding layer and then concatenated with other dense features as input to DCN-v2. The cross-network consists of 2 layers, with two hidden layers of the MLP producing outputs sized at 128 and 64 respectively. To further capture the link sequence attribute information of each route, the link sequence features are converted into high-dimensional features through the embedding layer. Then the features are projected into a 128-dimensional space by a fully connected layer with ReLU as the activation function. The transformed features are inputted into LSTM with a cell size of 256. Finally, the output of DCN-v2 and the last hidden state of the LSTM are combined, and the final score is generated via a sigmoid activation applied through a linear layer.

The embedding dimension of the link IDs is set to 32. Adam is used as the optimizer and the learning rate of Adam is set to 0.0001. The loss function is the cross-entropy loss defined below: 
\begin{equation}
  L=-\frac{1}{N} \sum_{i=1}^N \left[ y_i \log(\hat{y}_i) + (1 - y_i) \log(1 - \hat{y}_i) \right]
\end{equation}
where \(N\) is the size of the training set, \(y_i\) is the binary target, \(\hat{y}_i\) is the predicted value.

\section{Experiments}

\subsection{Datasets}

The dataset utilized in this study is sourced from the dataset of private car users who employ navigation services in our company. It encompasses the historical trajectory data and navigation records of 10,000 users over one month, covering cities such as Beijing, Shanghai, Guangzhou, and Shenzhen. The dataset comprises approximately 290,000 trajectories, with approximately 3 million candidate routes provided by the navigation service. Table 1 lists the statistics of the datasets.

\begin{table}[h]
  \caption{Statistics of datasets.}
  \label{tab:freq}
  \begin{tabular}{cccc}
    \toprule
     & number of links & navigation number & Average number of candidate routes\\
    \midrule
    training set & 2,105,063 & 236,660 & 11\\
    validation set & 1,369,689 & 26,295 & 11\\
    test set & 1,335,577 & 25,604 & 11\\
  \bottomrule
\end{tabular}
\end{table}

\subsection{Evaluation Metrics}

We use two metrics in our experiments, including the mean of inconsistency rate (mean\_IR) and AUC, to evaluate the performance of the method. The calculation formula of AUC is defined as follows:
\begin{equation}
  AUC=\frac{\sum I(p_{\text{pos}}, p_{\text{neg}})}{P \cdot N}
\end{equation}
\begin{equation}
  I(p_{\text{pos}}, p_{\text{neg}}) =
    \begin{cases} 
    1 & \text{if } p_{\text{pos}} > p_{\text{neg}} \\
    0.5 & \text{if } p_{\text{pos}} = p_{\text{neg}} \\
    0 & \text{if } p_{\text{pos}} < p_{\text{neg}}
    \end{cases}
\end{equation}
where \(P\) is the number of positive samples, \(N\) is the number of negative samples, \(p_{\text{pos}}\) represents the probability that the positive sample prediction is a positive example, \(p_{\text{neg}}\) represents the probability that the negative sample prediction is a negative example, and \(I\) is the indicator function.

\subsection{Competing method}

To validate the efficacy of the method proposed in this paper, we compared multiple solutions, including the minimum estimated time of arrival (ETA) solution, LightGBM \cite{Ke17}, and DCN-v2 model \cite{Wang21}. The minimum ETA solution indicates the recommended route with the fastest ETA. LightGBM solution indicates the prediction inconsistency rate based on the lightGBM model, and the input features do not include link sequence information. The input data of DCN-v2 is the same as that of LightGBM, only the model frame is different.

\subsection{Comparison results}

The DCR model effectively learns route information, sequential link information, and user preferences. The experimental results are presented in Table 2. It can be observed that compared to the minimum ETA solution, the mean\_IR of the LightGBM, DCN-v2, and DCR increased by 6.53\%, 7.82\%, and 8.72\% respectively. This improvement signifies that models leveraging large-scale historical data effectively capture routing recommendation information. The minimum ETA solution does not fully leverage the available information in the history data, thus its performance is unsatisfactory. 

\begin{table}[h]
  \caption{The result on the test dataset.}
  \label{tab:result}
  \begin{tabular}{lccc}
    \toprule
     & Test size & AUC & mean\_IR \\
    \midrule
    Min\_ETA & 25{,}604 & - & 38.97\%\\
    LightGBM & 25{,}604 & 81.09\% & 32.44\%\\
    DCN-v2 & 25{,}604 & 85.18\% & 31.15\%\\
    DCR & 25{,}604 & 86.32\% & 30.25\%\\
  \bottomrule
\end{tabular}
\end{table}

The DCR model was compared with the LightGBM and DCN-v2 models to assess the influence of modeling link sequence information along the route, in which input of LightGBM and DCN-v2 differ from DCR by removing the link sequence information. The forecasting results indicated that compared with LightGBM and DCN-v2 models, the AUC of DCR in the test set is increased by 5.23\% and 1.14\%, and mean\_IR is reduced by 2.19\% and 0.9\%, which confirms the benefit of introducing the recurrent network structure.
\begin{figure}[ht]
  \centering
  \includegraphics[width=0.5\linewidth]{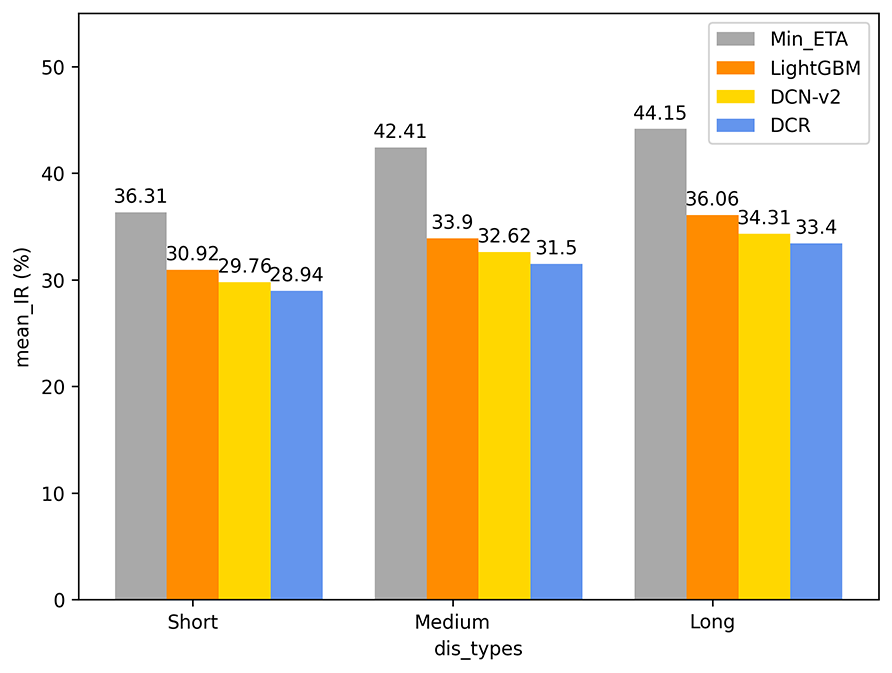}
  \caption{The result comparison of different distance types. The test dataset is classified into three distance types, and then the effects of different models on each set of data are compared.}
  \Description{}
\end{figure}

In addition to presenting the overall performance of the test dataset, we classify the test dataset into three distance types: short (0 to 10 km), medium (10 to 20 km), and long (longer than 20 km). Specifically, the test dataset comprises 15k short trajectories, 6k medium trajectories, and 4k long trajectories. As shown in Fig. 5, the performance is worse as the trajectory length increases, which indicates that shorter trips are more effective in predicting personalized routes for users. Additionally, across all distance types, the DCR model consistently outperforms other models significantly.

\section{Conclusion}

In this paper, we introduce a novel approach to learning personalized route recommendations. The goal is to learn the route ranking score and recommend the route with the best score to the user. To establish a comprehensive feature database, we construct route features encompassing temporal, spatial, traffic, and landscape aspects.  We use K-Means and T-SNE methods to extract user characteristics and effectively capture users’ behavioral preferences. Moreover, we propose a new deep learning model named DCR, which integrates DCN-v2 and LSTM to tackle the challenge. This model effectively learns user preferences and route information of route recommendations. We evaluated our approach offline with user navigation data and found that across the entire test datasets, the mean\_IR of the DCR was reduced by 8.72\% compared to the minimum ETA recommendation. When compared with LightGBM and DCN-v2, the DCR showed a reduction in mean\_IR by 2.19\% and 0.9\% respectively, and an increase in AUC by 5.23\% and 1.14\%. We categorized the test datasets into three groups based on short, medium, and long distances. As the distance increased, the mean\_IR decreased, indicating greater difficulty in predicting personalized routes for longer routes compared to shorter ones. Furthermore, the DCR outperformed other models in these evaluations. The result demonstrates the effectiveness of the approach.

\bibliographystyle{ACM-Reference-Format}
\bibliography{paper-references}


\begin{thebibliography}{10}


\ifx \showCODEN    \undefined \def \showCODEN     #1{\unskip}     \fi
\ifx \showDOI      \undefined \def \showDOI       #1{#1}\fi
\ifx \showISBNx    \undefined \def \showISBNx     #1{\unskip}     \fi
\ifx \showISBNxiii \undefined \def \showISBNxiii  #1{\unskip}     \fi
\ifx \showISSN     \undefined \def \showISSN      #1{\unskip}     \fi
\ifx \showLCCN     \undefined \def \showLCCN      #1{\unskip}     \fi
\ifx \shownote     \undefined \def \shownote      #1{#1}          \fi
\ifx \showarticletitle \undefined \def \showarticletitle #1{#1}   \fi
\ifx \showURL      \undefined \def \showURL       {\relax}        \fi
\providecommand\bibfield[2]{#2}
\providecommand\bibinfo[2]{#2}
\providecommand\natexlab[1]{#1}
\providecommand\showeprint[2][]{arXiv:#2}

\bibitem[Luo et~al\mbox{.}(2013)]%
        {Luo13}
\bibfield{author}{\bibinfo{person}{Wuman Luo}, \bibinfo{person}{Haoyu Tan}, \bibinfo{person}{Lei Chen}, {and} \bibinfo{person}{Lionel~M. Ni}.} \bibinfo{year}{2013}\natexlab{}.
\newblock \showarticletitle{Finding time period-based most frequent path in big trajectory data}. In \bibinfo{booktitle}{\emph{Proceedings of the 2013 ACM SIGMOD International Conference on Management of Data}} \emph{(\bibinfo{series}{SIGMOD '13})}. \bibinfo{publisher}{Association for Computing Machinery}, \bibinfo{address}{New York, NY}, \bibinfo{pages}{713–724}.
\newblock
\urldef\tempurl%
\url{https://doi.org/10.1145/2463676.2465287}
\showDOI{\tempurl}


\bibitem[Li et~al\mbox{.}(2022)]%
        {Li22}
\bibfield{author}{\bibinfo{person}{Peisong Li}, \bibinfo{person}{Xinheng Wang}, \bibinfo{person}{Honghao Gao}, \bibinfo{person}{Xiaolong Xu}, \bibinfo{person}{Muddesar lqbal}, {and} \bibinfo{person}{Keshav Dahal}.} \bibinfo{year}{2022}\natexlab{}.
\newblock \showarticletitle{A Dynamic and Scalable User-Centric Route Planning Algorithm Based on Polychromatic Sets Theory}.
\newblock \bibinfo{journal}{\emph{IEEE Transactions on Intelligent Transportation Systems}} \bibinfo{volume}{23}, \bibinfo{number}{3} (\bibinfo{year}{2022}), \bibinfo{pages}{2762--2772}.
\newblock
\urldef\tempurl%
\url{https://doi.org/10.1109/TITS.2021.3085026}
\showDOI{\tempurl}


\bibitem[Wang et~al\mbox{.}(2019)]%
        {Wang19}
\bibfield{author}{\bibinfo{person}{Jingyuan Wang}, \bibinfo{person}{Ning Wu}, \bibinfo{person}{Wayne~Xin Zhao}, \bibinfo{person}{Fanzhang Peng}, {and} \bibinfo{person}{Xin Lin}.} \bibinfo{year}{2019}\natexlab{}.
\newblock \showarticletitle{Empowering A* Search Algorithms with Neural Networks for Personalized Route Recommendation}. In \bibinfo{booktitle}{\emph{Proceedings of the 25th ACM SIGKDD International Conference on Knowledge Discovery \& Data Mining}} \emph{(\bibinfo{series}{KDD '19})}. \bibinfo{publisher}{Association for Computing Machinery}, \bibinfo{address}{New York, NY}, \bibinfo{pages}{539–547}.
\newblock
\urldef\tempurl%
\url{https://doi.org/10.1145/3292500.3330824}
\showDOI{\tempurl}


\bibitem[Cheng et~al\mbox{.}(2021)]%
        {Cheng21}
\bibfield{author}{\bibinfo{person}{Ran Cheng}, \bibinfo{person}{Chao Chen}, \bibinfo{person}{Longfei Xu}, \bibinfo{person}{Shen Li}, \bibinfo{person}{Lei Wang}, \bibinfo{person}{Hengbin Cui}, \bibinfo{person}{Kaikui Liu}, {and} \bibinfo{person}{Xiaolong Li}.} \bibinfo{year}{2021}\natexlab{}.
\newblock \bibinfo{title}{R4: A Framework for Route Representation and Route Recommendation}.
\newblock
\newblock
\urldef\tempurl%
\url{https://doi.org/10.48550/arXiv.2110.10474}
\showURL{%
\tempurl}


\bibitem[Liu et~al\mbox{.}(2020)]%
        {Liu20}
\bibfield{author}{\bibinfo{person}{Shan Liu}, \bibinfo{person}{Hai Jiang}, \bibinfo{person}{Shuiping Chen}, \bibinfo{person}{Jing Ye}, \bibinfo{person}{Renqing He}, {and} \bibinfo{person}{Zhizhao Sun}.} \bibinfo{year}{2020}\natexlab{}.
\newblock \showarticletitle{Integrating Dijkstra’s algorithm into deep inverse reinforcement learning for food delivery route planning}.
\newblock \bibinfo{journal}{\emph{Transportation Research Part E: Logistics and Transportation Review}}  \bibinfo{volume}{142} (\bibinfo{year}{2020}), \bibinfo{pages}{102070}.
\newblock
\urldef\tempurl%
\url{https://doi.org/10.1016/j.tre.2020.102070}
\showDOI{\tempurl}


\bibitem[Liu and Jiang(2022)]%
        {Liu22}
\bibfield{author}{\bibinfo{person}{Shan Liu} {and} \bibinfo{person}{Hai Jiang}.} \bibinfo{year}{2022}\natexlab{}.
\newblock \showarticletitle{Personalized route recommendation for ride-hailing with deep inverse reinforcement learning and real-time traffic conditions}.
\newblock \bibinfo{journal}{\emph{Transportation Research Part E: Logistics and Transportation Review}}  \bibinfo{volume}{164} (\bibinfo{year}{2022}), \bibinfo{pages}{102780}.
\newblock
\urldef\tempurl%
\url{https://doi.org/10.1016/j.tre.2022.102780}
\showDOI{\tempurl}


\bibitem[der Maaten and Hinton(2008)]%
        {Maaten08}
\bibfield{author}{\bibinfo{person}{Laurens~Van der Maaten} {and} \bibinfo{person}{Geoffrey Hinton}.} \bibinfo{year}{2008}\natexlab{}.
\newblock \showarticletitle{Visualizing data using t-SNE}.
\newblock \bibinfo{journal}{\emph{Journal of Machine Learning Research}} \bibinfo{volume}{9}, \bibinfo{number}{86} (\bibinfo{year}{2008}), \bibinfo{pages}{2579--2605}.
\newblock


\bibitem[Wang et~al\mbox{.}(2021)]%
        {Wang21}
\bibfield{author}{\bibinfo{person}{Ruoxi Wang}, \bibinfo{person}{Rakesh Shivanna}, \bibinfo{person}{Derek Cheng}, \bibinfo{person}{Sagar Jain}, \bibinfo{person}{Dong Lin}, \bibinfo{person}{Lichan Hong}, {and} \bibinfo{person}{Ed Chi}.} \bibinfo{year}{2021}\natexlab{}.
\newblock \showarticletitle{DCN V2: Improved Deep \& Cross Network and Practical Lessons for Web-scale Learning to Rank Systems}. In \bibinfo{booktitle}{\emph{Proceedings of the Web Conference 2021}} \emph{(\bibinfo{series}{WWW '21})}. \bibinfo{publisher}{Association for Computing Machinery}, \bibinfo{address}{New York, NY}, \bibinfo{pages}{1785–1797}.
\newblock
\urldef\tempurl%
\url{https://doi.org/10.1145/3442381.3450078}
\showDOI{\tempurl}


\bibitem[Hochreiter and Schmidhuber(1997)]%
        {Hochreiter1997}
\bibfield{author}{\bibinfo{person}{Sepp Hochreiter} {and} \bibinfo{person}{Jürgen Schmidhuber}.} \bibinfo{year}{1997}\natexlab{}.
\newblock \showarticletitle{Long short-term memory}.
\newblock \bibinfo{journal}{\emph{Neural Computation}} \bibinfo{volume}{9}, \bibinfo{number}{8} (\bibinfo{year}{1997}), \bibinfo{pages}{1735–1780}.
\newblock


\bibitem[Ke et~al\mbox{.}(2017)]%
        {Ke17}
\bibfield{author}{\bibinfo{person}{Guolin Ke}, \bibinfo{person}{Qi Meng}, \bibinfo{person}{Thomas Finley}, \bibinfo{person}{Taifeng Wang}, \bibinfo{person}{Wei Chen}, \bibinfo{person}{Weidong Ma}, \bibinfo{person}{Qiwei Ye}, {and} \bibinfo{person}{Tie-Yan Liu}.} \bibinfo{year}{2017}\natexlab{}.
\newblock \showarticletitle{LightGBM: a highly efficient gradient boosting decision tree}. In \bibinfo{booktitle}{\emph{Proceedings of the 31st International Conference on Neural Information Processing Systems}} \emph{(\bibinfo{series}{NIPS'17})}. \bibinfo{publisher}{Curran Associates Inc.}, \bibinfo{address}{Red Hook, NY}, \bibinfo{pages}{3149–3157}.
\newblock


\end{thebibliography}

\appendix

\end{document}